\pdfoutput=1

\documentclass[11pt]{article}

\usepackage[final]{acl}

\usepackage{times}
\usepackage{latexsym}

\usepackage[T1]{fontenc}

\usepackage[utf8]{inputenc}

\usepackage{microtype}

\usepackage{inconsolata}

\usepackage{graphicx}

\newcommand{\stitle}[1]{\vspace{3pt} \noindent\textbf{#1}}

\usepackage{color}
\definecolor{orange}{rgb}{1,0.647,0}

\usepackage{graphicx}
\usepackage{hyperref}
\usepackage{url}
\usepackage{xcolor}
\usepackage{tabularx}
\usepackage{multicol}
\usepackage{colortbl}
\usepackage{booktabs}
\usepackage{xspace}
\usepackage{algorithm}
\usepackage{algpseudocode}
\usepackage{bbm, mathtools}
\usepackage{wrapfig}
\usepackage{cuted}
\usepackage{hyperref}

\usepackage{tcolorbox}

\def\Appref#1{Appx.~\ref{#1}}

\def\Tabref#1{Tab.~\ref{#1}}

\newcommand{\method}{\mbox{\textsc{Astute RAG}}\xspace}

\usepackage{amsmath,amsfonts,bm}

\def\Figref#1{Fig.~\ref{#1}}

\def\Secref#1{Sec.~\ref{#1}}

\def\eqref#1{equation~\ref{#1}}

\def\Algref#1{Alg.~\ref{#1}}

\def\1{\bm{1}}

\DeclareMathAlphabet{\mathsfit}{\encodingdefault}{\sfdefault}{m}{sl}
\SetMathAlphabet{\mathsfit}{bold}{\encodingdefault}{\sfdefault}{bx}{n}

\title{\method: Overcoming Imperfect Retrieval Augmentation and Knowledge Conflicts for Large Language Models}

\author{
Fei Wang$^{12*}$ \quad 
Xingchen Wan$^{1}$ \quad 
Ruoxi Sun$^{1}$ \quad 
Jiefeng Chen$^{1}$ \quad 
Sercan Ö. Arık$^{1}$ \quad 
\\
$^1$Google Cloud AI Research \quad 
$^2$University of Southern California
\\
\small \texttt{fwang598@usc.edu} \quad 
\texttt{\{xingchenw,ruoxis,jiefengc,soarik\}@google.com}
}

\begin{document}
\maketitle

\renewcommand{\thefootnote}{\fnsymbol{footnote}}
\footnotetext[1]{Work done during internship at Google.}
\renewcommand{\thefootnote}{\arabic{footnote}}

\begin{abstract}

Retrieval-augmented generation (RAG), while effective in integrating external knowledge to enhance large language models (LLMs), can be undermined by \textit{imperfect} retrieval, which may introduce irrelevant, misleading, or even malicious information. Despite its importance, previous studies have rarely explored the behavior of RAG with errors from imperfect retrieval, and how potential conflicts arise between the LLMs' internal knowledge and external sources. We show that imperfect retrieval augmentation might be inevitable and quite harmful, through controlled analysis under realistic conditions. \textit{Knowledge conflicts} between LLM-internal and external knowledge from retrieval is a bottleneck to overcome in the post-retrieval stage of RAG. To render LLMs resilient to imperfect retrieval, we propose \method, a novel RAG approach that \textit{adaptively} elicits essential information from LLMs' internal knowledge, \textit{iteratively} consolidates internal and external knowledge with \textit{source-awareness}, and finalizes the answer according to information reliability. Our experiments with Gemini and Claude demonstrate that \method significantly outperforms previous robustness-enhanced RAG methods. Notably, \method is the only approach that matches or exceeds the performance of LLMs without RAG under worst-case scenarios. \method effectively resolves knowledge conflicts, improving the reliability and trustworthiness of RAG systems.
\end{abstract}

\section{Introduction}

\begin{figure*}[t] 
\centering
\includegraphics[width=\textwidth]{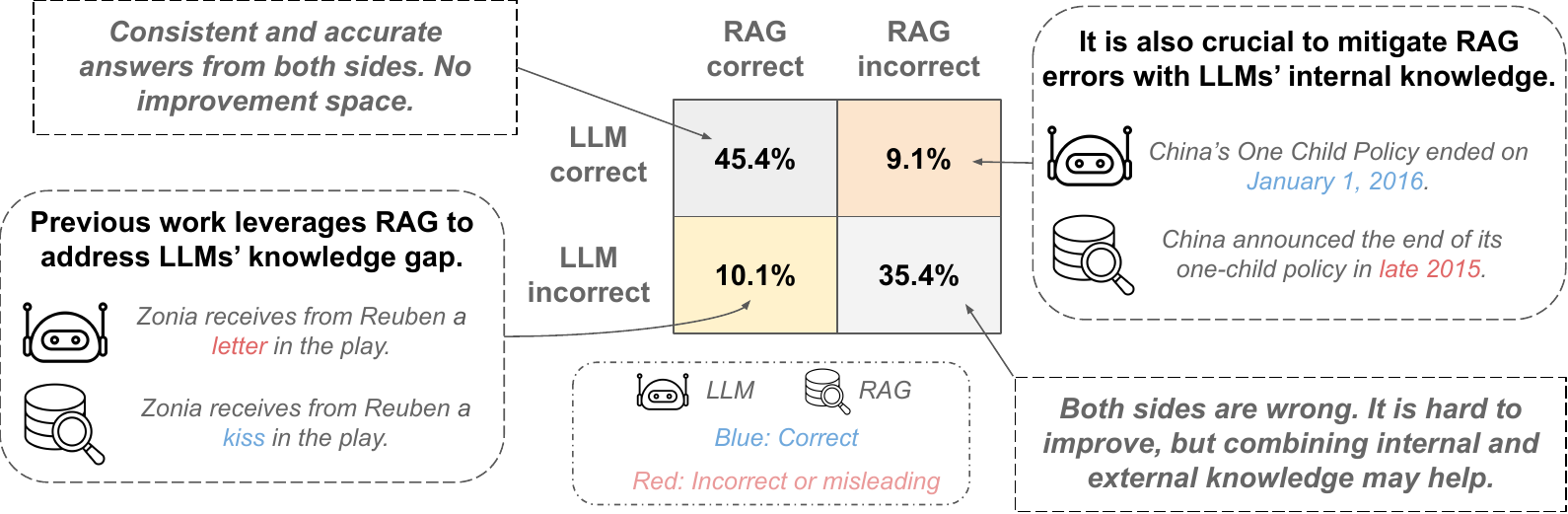}
\caption{Knowledge conflicts between the LLMs' internal knowledge and retrieved knowledge from external sources. We report the overall results with Claude under the setting in \Secref{sec:experiment_setting}.}
\label{fig:problem}
\end{figure*}

Retrieval-augmented generation (RAG) is commonly used for large language models (LLMs) to tackle knowledge-intensive tasks \citep{guu2020retrieval,lewis2020retrieval}. 
Prior works mainly leverage RAG to address the inherent knowledge limitations of LLMs, effectively integrating missing information and grounding to reliable sources.
However, recent research has highlighted a significant drawback that RAG might rely on \textit{imperfect retrieval}, including irrelevant, misleading, or even malicious information (\Figref{fig:problem}), which eventually leads to inaccurate LLM responses \citep{chen2024benchmarking,zou2024poisonedrag}. 
Moreover, recent studies have shown that retrieval augmentation can confuse LLMs when retrieved passages are \textit{conflicting} with LLMs' parametric knowledge \cite{tan2024blinded,xie2024adaptive,jin2024tug}.
These pose significant challenges to the trustworthiness of RAG. %

To address imperfect retrieval, earlier work seeks to improve the retrieval approaches, such as dynamic and iterative retrieval \cite{jiang2023active,asai2023self,yan2024corrective}.
However, the occurrence of imperfect retrieval is still inevitable, due to corpus quality limitations \citep{shao2024scaling}, the reliability of retrievers \citep{dai2024unifying}, and the complexity of queries \citep{su2024bright}. 
Consequently, recent work shifts the focus to the generation stage, seeking to reduce the negative impact of noisy retrieved passages~\cite{xiang2024certifiably,wei2024instructrag}.
Another line of research at generation stage, motivated by knowledge conflicts, has explored complementing retrieved passages with LLM-generated passages \cite{yu2023generate,zhang2023merging} or deactivating RAG when the retrieved passages are of insufficient quality~\cite{xu2024retrieval,mallen2023not,jeong2024adaptive}.

Despite the previous work on the impact of imperfect retrieval and knowledge conflicts at RAG generation stage, quantitative analyses lack on two crucial real-world aspects:
(i) the relation between retrieval quality and occurrence of knowledge conflicts, 
and (ii) the extent to which retrieved passages and LLMs' parametric knowledge can correct each other. Method-wise, existing approaches for mitigating RAG failures caused by imperfect retrieval and knowledge conflicts have not yet yielded a training-free method capable of \textit{explicitly} analyzing conflicting knowledge across various internal and external sources, and achieving \textit{worst-case} robustness for black-box LLMs.

In this paper, we first conduct comprehensive analyses to investigate the relation between imperfect retrieval and knowledge conflicts, and examine the frequency of external and LLMs' internal knowledge mutually correcting each other~(\Secref{sec:benchmark}). 
On a diverse range of general, domain-specific, and long-tail questions from NQ~\citep{kwiatkowski2019natural}, TriviaQA \citep{joshi2017triviaqa}, BioASQ \citep{tsatsaronis2015overview}, and PopQA \citep{mallen2023not}, we observe that imperfect retrieval is widespread even with an adept real-world search engine, leading to the impeded performance of RAG.\footnote{such as Google Search with Web as corpus} 
Retrieval precision is tightly correlated with the knowledge conflict rate.
Mutual correction between the LLM's knowledge and external knowledge is crucial for recovering from RAG failures.
Our findings underscore the potential severity of imperfect retrieval in real-world RAG and highlight the widespread existence of knowledge conflicts as the bottleneck. 

We propose \method, a novel RAG approach designed for resilience to imperfect retrieval augmentation, while preserving RAG grounding effect when retrieval is reliable (\Secref{sec:method}). \method effectively differentiates between consistent and conflicting information from the LLM's internal knowledge and the externally retrieved passages, assesses their reliability, and ensures proper integration of trustworthy information. \method first adaptively elicits LLMs' knowledge and then conducts source-aware knowledge consolidation. The desiderata is combining consistent information, identifying conflicting information, and filtering out irrelevant information. Finally, \method proposes answers based on consistent information and compares them to determine the final answer.
Our experiments with various LLMs (Claude, Gemini and Mistral), demonstrate superior performance of \method compared to previous RAG approaches designed for robustness~(\Secref{sec:experiment}). 
\method consistently outperforms baselines across different retrieval quality levels. 
Notably, \method is the only RAG method that achieves performance comparable to or even surpassing retrieval-free mode of LLMs under the worst-case scenario where all retrieved passages are unhelpful. 
Further analysis reveals the effectiveness of \method in resolving knowledge conflicts.

In summary, our core contributions are threefold. 
First, we provide quantitative analyses and novel insights for the connection among imperfect retrieval, knowledge conflicts, and RAG failures under real-world conditions. 
Second, we propose \method, which explicitly analyzes LLM-internal and external knowledge in-context, assesses their reliablity, and recovers from RAG failures with black-box access.
Third, with experiments with various LLMs and datasets, we demonstrate the effectiveness of \method in improving robustness and trustworthiness, even in the most challenging scenarios.

\section{Related Work}

\begin{figure*}[t]
\centering
\includegraphics[width=\textwidth]{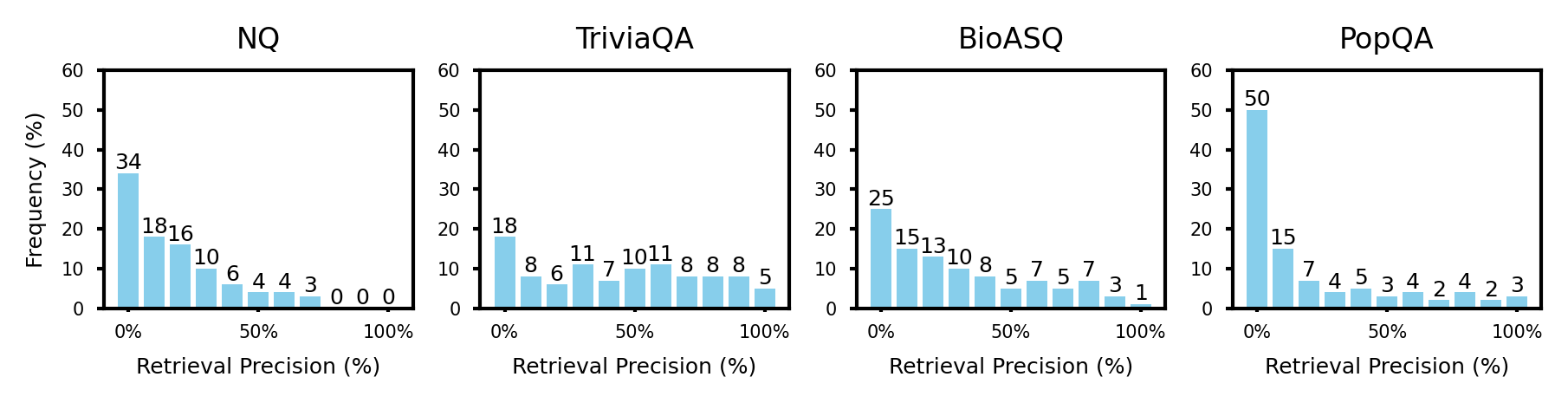}
\caption{Imperfect retrieval (samples with low retrieval precision) is prevalent in real-world RAG.}
\label{fig:distribution}
\end{figure*}

\begin{figure}[t]
\centering
\includegraphics[width=\columnwidth]{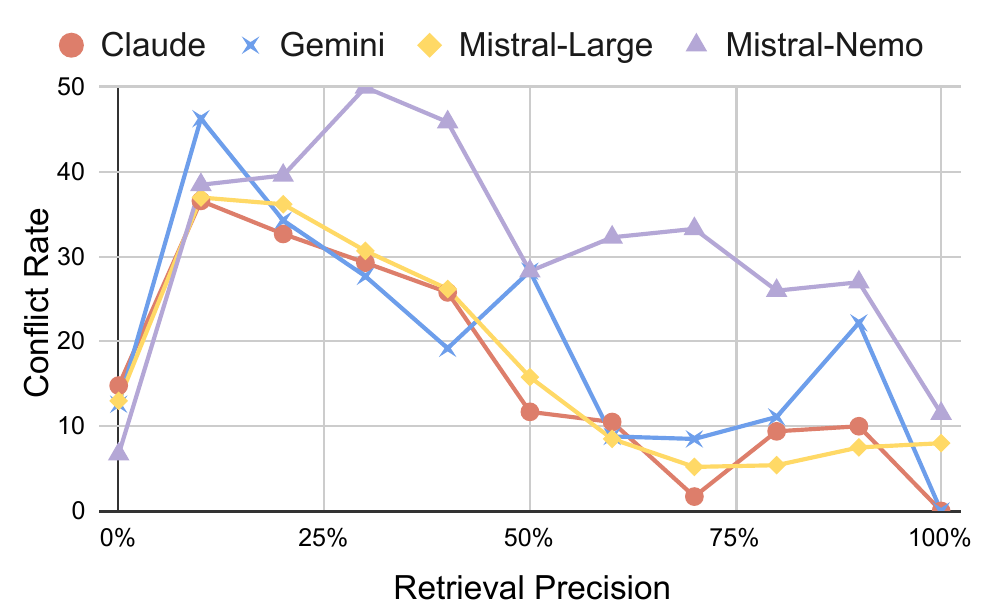}
\caption{Conflicting rate between answers from LLMs with and without RAG on different retrieval precision.}
\label{fig:conflict_rate}
\end{figure}

RAG aims to address the inherent knowledge limitation of LLMs with passages retrieved from external sources of information such as private corpora or public knowledge bases \citep{guu2020retrieval,lewis2020retrieval,borgeaud2022improving}. 
Given the widespread real-world adoption of RAG, including risk-sensitive domains, the negative impact of noisy information within retrieved passages has garnered increasing attention \citep{cuconasu2024power}.
Recent work explored enhancing the robustness of RAG systems against noise from various perspectives, including training LLMs with noisy context  \citep{yu2023chain,yoran2024making,pan2024not,fang2024enhancing}, training small models to filter out irrelevant passages \citep{wang2023learning,xu2023recomp}, passage reranking \citep{yu2024rankrag,glass2022re2g}, dynamic and iterative retrieval \citep{jiang2023active,asai2023self,yan2024corrective}, query rewriting \citep{ma2023query}, and speculative drafting \citep{wang2024speculative}. These focus on distinct modules or stages of RAG systems and are orthogonal to our work.

Our work focuses on enhancing RAG robustness at the post-retrieval stage, after retrieved passages have been provided. On this, RobustRAG \citep{xiang2024certifiably} aggregates answers from each independent passage to provide certifiable robustness. InstructRAG \citep{wei2024instructrag} instructs the LLM to provide a rationale connecting the answer with information in passages. MADRA \citep{wang2023apollo} applies multi-agent debate to select helpful evidence. However, these do not explicitly incorporate internal knowledge to recover from RAG failures and therefore might severely suffer when the majority of retrieved passages have issues.
For emphasizing internal knowledge of LLMs in RAG, recent work explored using LLM-generated passage as context \citep{yu2023generate}, training models to match generated and retrieved passages \citep{zhang2023merging}, adaptively switching between LLMs with and without RAG \citep{xu2024retrieval,mallen2023not,jeong2024adaptive}, and combining answers through contrastive decoding  \citep{zhao2024enhancing,jin2024tug}. 
Different from prior work, we provide a systematic framework on connecting imperfect retrieval, knowledge conflicts, and RAG failures.
Specifically focusing on the imperfect context setting, our method is training-free and applicable to black-box LLMs, explicitly analyzes internal and external knowledge in-context, and offers broader usability and adaptability.

\section{The Pitfall of RAG}
\label{sec:benchmark}

To better showcase common real-world challenges and motivate improved methodological designs, we evaluate retrieval quality, the occurrence of knowledge conflicts, their relationship, and the mutual correction between external and internal knowledge using a controlled dataset derived from NQ, TriviaQA, BioASQ, and PopQA, datasets widely used for RAG in prior work \cite{xiang2024certifiably,wei2024instructrag,asai2023self}.
Different from prior work, our analysis is based on real-world retrieval results with Google Search\footnote{\url{https://developers.google.com/custom-search/v1/overview}} as the retriever and the Web as the corpus. 
Overall, we sample 1K instances, each with 10 retrieved passages.

\begin{figure*}[t] 
\centering
\includegraphics[width=\textwidth]{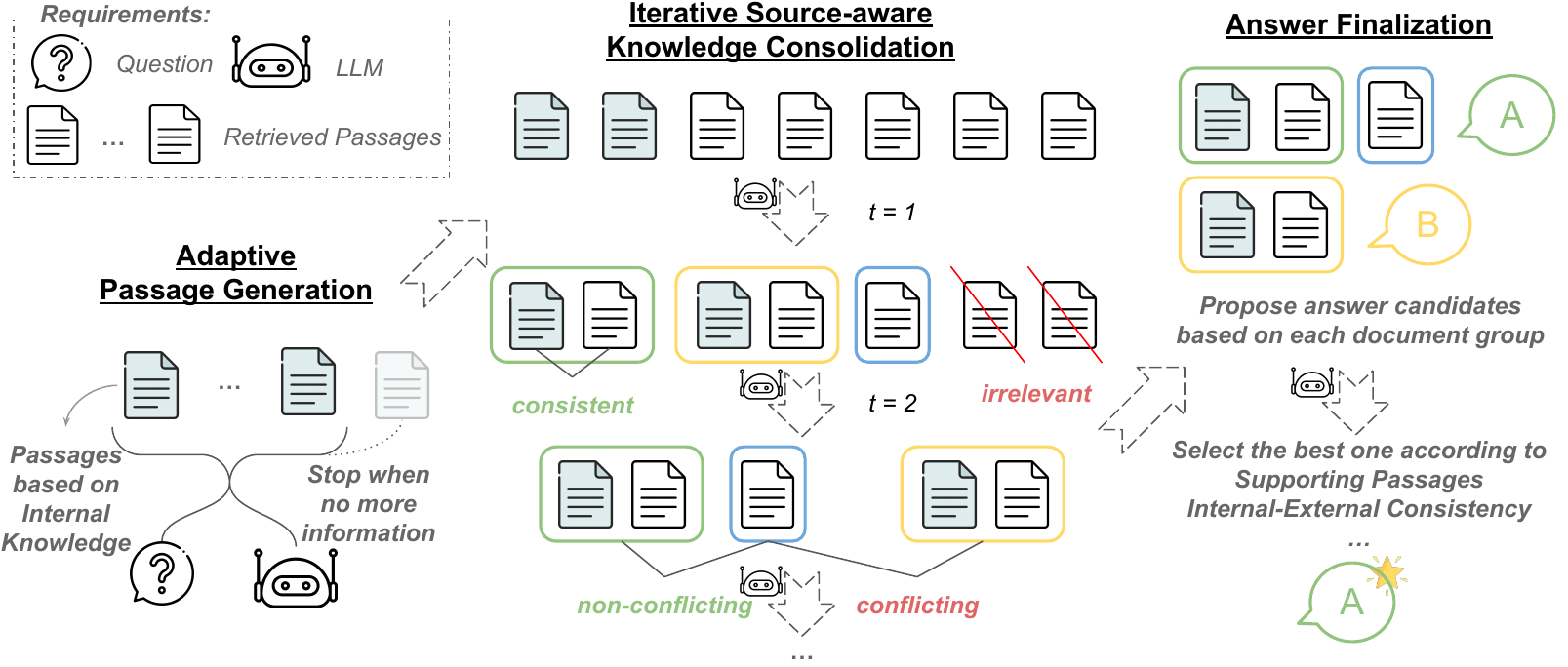}
\caption{Overview of the \method framework. \method is designed to better combine the information from the external sources (e.g. web, domain-specific corpora) and internal knowledge of the LLMs by employing a consolidation mechanism to address the conflicts, which eventually leads to superior generation quality.}
\label{fig:method}
\end{figure*}

\stitle{Imperfect retrieval and knowledge conflicts are common and harmful.}
Our initial observations are consist with prior work.
As shown in \Figref{fig:distribution}, the retrieval precision\footnote{Ratio of passages directly contain true answers.} is generally low - roughly 70\% retrieved passages do not directly contain true answers, consistent with prior work demonstrating the often imperfect nature of retrieval results \cite{thakur2beir,su2024bright}.
With Claude 3.5 Sonnet as the LLM, \Figref{fig:problem} shows that 19.2\% of the overall data exhibit knowledge conflicts, consistent with prior work demonstrating the prevalence of such conflicts across various scenarios \cite{pham2024s,xie2024adaptive,longpre2021entity}.
Moreover, we observe strong correlations between retrieval precision and RAG performance (\Figref{fig:perf_by_retrieval}) and between the occurrence of knowledge conflicts and RAG performance (\Figref{fig:perf_by_conflicts}), findings consistent with prior work on these respective topics \cite{chen2024benchmarking,xie2024adaptive}.

\stitle{Lower retrieval precision increases knowledge conflicts in general.}
As shown in \Figref{fig:conflict_rate}, most advanced LLMs exhibit the highest conflict rates when retrieval precision is as low as 10\%. Subsequently, the conflict rate generally decreases as precision increases, although some fluctuations may occur. This trend is generally applicable to the studied LLMs with different training processes. Notably, when retrieval precision is 0\%, conflict rates tend to be significantly lower. This suggests that limited external knowledge for the query results in more irrelevant passages rather than incorrect ones.

\stitle{Internal and external knowledge can correct each other to a comparable extent.}
Among the conflicting cases, the internal knowledge is correct on 47.4\% of them, while the external knowledge is correct on the remaining 52.6\%. These results emphasize the importance of effectively combining the internal and external knowledge to overcome the inherent limitation of relying solely on either source. However, previous work \citep{tan2024blinded,xie2024adaptive,jin2024tug} shows that LLMs often select knowledge based on unreliable shortcuts, so simply presenting LLM-generated passages in the context may not help.

\section{\method}  %
\label{sec:method}

\begin{algorithm*}[t]
\small
\caption{\method}\label{alg:method}
\begin{algorithmic}[1] 
\Require  Query $q$,  Retrieved Passages $E=[e_1, \ldots, e_n]$, Large Language Model $\mathcal{M}$, Number of Iteration $t$, Max Number of Generated Passages $\hat{m}$, Prompt Templates $p_{gen}, p_{con}, p_{ans}$
\State  Adaptively generate passages:  $I \leftarrow \mathcal{M}(p_{gen}, q, \hat{m})$  \Comment{\textcolor{blue}{\Secref{sec:method_internal}}}
\State Combine internal and external passages: $D_{0} \leftarrow E \oplus I$
\State Assign passage sources: $S_{0} \leftarrow [\mathbbm{1}_{\{d \in E\}} \text{for} \ d  \ \text{in} \ D_0]$
\If{$t > 1$}
    \For{$j = 1, \ldots, t-1$} \Comment{\textcolor{blue}{\Secref{sec:method_consolidation}}}
        \State Consolidate knowledge: $\langle D_{j+1}, S_{j+1} \rangle  \leftarrow \mathcal{M}(p_{con}, q, \langle D_{0}, S_{0} \rangle, \langle D_{j}, S_{j} \rangle)$ 
    \EndFor
    \State Finally consolidate and answer: $a \leftarrow \mathcal{M}(p_{ans}, q, \langle D_{0}, S_{0} \rangle, \langle D_{t-1}, S_{t-1} \rangle)$  \Comment{\textcolor{blue}{\Secref{sec:method_answer}}}
\Else
    \State Consolidate knowledge and finalize the answer: $a \leftarrow \mathcal{M}(p_{ans}, q,\langle D_0, S_0 \rangle)$
\EndIf
\State \Return $a$
\end{algorithmic}
\end{algorithm*}

We first provide an overview of \method (\Secref{sec:method_overview}). Subsequently, we delve into the three major steps of \method, including adaptive generation of internal knowledge (\Secref{sec:method_internal}), source-aware knowledge consolidation (\Secref{sec:method_consolidation}), and answer finalization (\Secref{sec:method_answer}).

\subsection{Overview}
\label{sec:method_problem}
\label{sec:method_overview}

Our objective is to mitigate the effects of imperfect retrieval augmentation, resolve knowledge conflicts between the LLM's internal knowledge and external sources (such as custom/public corpora and knowledge bases), and ultimately produce more accurate and reliable responses from LLMs.
Given a set of retrieved passages from external sources $E=[e_1, \ldots, e_n]$, a pre-trained LLM $\mathcal{M}$ (accessible through prediction-only APIs, encompassing commercial black-box ones), and a query $q$, the task is to generate the corresponding correct answer $a^{*}$.
Notably, this setting is orthogonal to prior work on improving the retriever, training LLMs, or conducting adaptive retrieval, which are mainly preliminary steps.

\method is designed to better leverage collective knowledge from both internal knowledge of LLMs and external corpus, for more reliable responses. As shown in \Figref{fig:method} and \Algref{alg:method}, \method starts from acquiring the most accurate, relevant, and thorough passage set from the LLMs' internal knowledge. Then, internal and external knowledge are consolidated in an iterative way, by comparing the generated and retrieved passages. Finally, the reliability of conflicting information is compared and the final output is generated according to the most reliable knowledge.

\subsection{Adaptive Generation of Internal Knowledge}
\label{sec:method_internal}
In the first step, we elicit internal knowledge from LLMs. This LLM-internal knowledge, reflecting the consensus from extensive pre-training and instruction-tuning data, can supplement any missing information from the limited set of retrieved passages and enable mutual confirmation between LLM-internal and external knowledge. This is especially valuable when the majority of retrieved passages might be irrelevant or misleading.
Specifically, we prompt LLMs to generate passages based on the given question $q$, following \citet{yu2023generate}. While \citet{yu2023generate} primarily focused on generating diverse internal passages, we emphasize the importance of reliability and trustworthiness of generated passages. To achieve this goal, we enhance the original method with \textit{constitutional principles} and \textit{adaptive generation}.

Inspired by \citet{bai2022constitutional}, we provide \textbf{constitutional principles} indicating the desired properties of internal passages in the prompt $p_{gen}$ (see \Appref{sec:appendix_prompt} for details) to guide their generation, emphasizing that the generated passages should be accurate, relevant, and hallucination-free. Moreover, we allow the LLM to perform \textbf{adaptive generation} of passages in its internal knowledge. The LLM can decide how many passages to generate by itself. Rather generating a fix number of passages, we request the LLM to generate at most $\hat m$ passages, each covering distinct information, and to directly indicate if no more reliable information is available. This adaptive approach allows the LLM to generate fewer passages (or even no passages at all) when the useful information within internal knowledge is limited and more passages when there are multiple feasible answers in the internal knowledge. In this step, the LLM generates $m \leq \hat{m}$ passages based on its internal knowledge:
\[
I = [i_{1}, \ldots i_{m}] = \mathcal{M}(p_{gen}, q, \hat{m}).
\]

\subsection{Iterative Source-aware Knowledge Consolidation}
\label{sec:method_consolidation}
In the second step, we employ the LLM to explicitly consolidate information from both passages generated from its internal knowledge and passages retrieved from external sources.
Initially, we combine passages from both internal and external knowledge sources $D_0 = E \oplus I.$

We additionally ensure \textbf{source-awareness} by providing the source of each passage to LLMs when consolidating knowledge. The source information (internal or external, such as a website) is helpful in assessing the reliability of passages. Here, we provide the passage source as
$S_0 = [\mathbbm{1}_{\{d \in E\}} \text{for} \ d  \ \text{in} \ D_0].$
To consolidate knowledge, we prompt the LLM (with $p_{con}$ in \Appref{sec:appendix_prompt}) to identify consistent information across passages, detect conflicting information between each group of consistent passages, and filter out irrelevant information. This step would regroup the unreliable knowledge in input passages into fewer refined passages.
The regrouped passages also attribute their source to the corresponding input passages:
$$\langle D_{j+1}, S_{j+1} \rangle  = \mathcal{M}(p_{con}, q,  \langle D_{0}, S_{0} \rangle, \langle D_{j}, S_{j} \rangle).$$
We find that this is especially helpful in comparing the reliability of conflicting knowledge and addressing knowledge conflicts. 
This knowledge consolidation process can run \textbf{iteratively} for $t$ times to improve better utilization of the retrieved context.

\subsection{Answer Finalization}
\label{sec:method_answer}
In the last step, we prompt the LLM (with $p_{ans}$ in \Appref{sec:appendix_prompt}) to generate one answer based on each group of passages ($\langle D_{t}, S_{t} \rangle$), and then compare their reliability and select the most reliable one as the final answer. This comparison allows the LLM to comprehensively consider knowledge source, cross-source confirmation, frequency, and information thoroughness when making the final decision. 
Notably, this step can be merged into the last knowledge consolidation step to reduce the inference complexity (the amount of prediction API calls) using a combined prompt:
$$a  = \mathcal{M}(p_{ans}, q, \langle D_{0}, S_{0} \rangle, \langle D_{t}, S_{t} \rangle).$$
When $t=1$, the initial passages will be input to the model directly for knowledge consolidation and subsequent answering: $a  = \mathcal{M}(p_{ans}, q, \langle D_{0}, S_{0} \rangle).$

\section{Experiments}
\label{sec:experiment}

\begin{table*}[t]
    \centering
    \small
    \setlength{\tabcolsep}{3pt} 
    \begin{tabular}{lcccc>{\columncolor{blue!5}}c|cccc>{\columncolor{blue!5}}c}
        \toprule
         Method & NQ & TriviaQA & BioASQ & PopQA &  Overall & NQ & TriviaQA & BioASQ & PopQA &  Overall \\ \midrule
        & \multicolumn{5}{c|}{\cellcolor{gray!25}\textit{Claude 3.5 Sonnet (20240620)}}  & \multicolumn{5}{c}{\cellcolor{gray!25}\textit{Gemini 1.5 Pro (002)}}    \\ \midrule
        No RAG & 47.1 & 82.0 & 50.4 & 29.8 & 54.5 & 44.8 & 80.2 & 45.8 & 25.3 & 51.3 \\ 
        RAG & 44.4 & 76.7 & 58.0 & 36.0 & 55.5 & 42.7 & 76.0 & 55.2 & 33.7 & 53.7 \\ \midrule
        USC \citep{chen2024universal} & 48.1 & 80.2 & \textbf{61.5} & 37.6 & 58.7 & 46.4 & 76.7 & \textbf{58.4} & 37.6 & 56.4 \\ \midrule
        GenRead \citep{yu2023generate} & 42.0 & 74.2 & 57.0 & 34.3 & 53.6 & 45.1 & 77.4 & 54.9 & 34.3 & 54.7 \\
        RobustRAG \citep{xiang2024certifiably} & 47.8 & 78.1 & 56.3 & 37.1 & 56.5 & 34.2 & 67.5 & 44.1 & 32.0 & 45.6\footnotemark \\
        InstructRAG \citep{wei2024instructrag} & 47.1 & 83.0 & 58.0 & 41.0 & 58.8 & 46.8 & 80.6 & 54.9 & 34.8 & 56.1 \\ 
        Self-Route \citep{xu2024retrieval} & 47.5 & 78.8 & 59.1 & 41.0 & 58.1 & 47.5 & 79.9 & 58.0 & 38.2 & 57.6 \\
        \midrule
        \method & \textbf{52.2} & \textbf{84.1} & 60.1 & \textbf{44.4} & \textbf{61.7} & \textbf{50.2} & \textbf{81.6} & 58.0 & \textbf{40.5} & \textbf{59.2}  \\ \midrule
        & \multicolumn{5}{c|}{\cellcolor{gray!25}\textit{Mistral-Large (2407), 128B}}  & \multicolumn{5}{c}{\cellcolor{gray!25}\textit{Mistral-Nemo (2407), 12B}}    \\ \midrule
        No RAG & 46.8 & 79.5 & 43.7 & 24.7 & 51.1 & 29.8 & 67.8 & 34.3 & 23.0 & 40.2 \\ 
        RAG & 43.1 & 77.4 & 55.9 & 36.0 & 54.7 & 39.3 & 66.8 & 49.0 & 32.6 & 48.3 \\ \midrule
        USC \citep{chen2024universal} & \textbf{51.2} & 80.9 & \textbf{61.5} & 36.0 & 59.5 & 29.5 & 66.1 & 36.0 & 20.2 & 39.6 \\ \midrule
        GenRead \citep{yu2023generate} & 40.7 & 73.1 & 55.6 & 35.4 & 52.7 & 38.6 & 68.9 & 48.3 & \textbf{33.7} & 48.7  \\ 
        RobustRAG \citep{xiang2024certifiably} & 42.7 & 77.7 & 50.4 & 34.8 & 53.0 & 35.6 & 71.7 & 44.1 & 27.5 & 46.4 \\ 
        InstructRAG \citep{wei2024instructrag} & 45.4 & 80.6 & 57.3 & 36.5 & 56.7 & 38.3 & 61.8 & 50.4 & 23.6 & 45.5 \\ 
        Self-Route \citep{xu2024retrieval} & 45.4 & 77.7 & 57.3 & 38.2 & 56.2 & 41.4 & 73.5 & \textbf{51.8} & 30.9 & 51.2 \\ \midrule
        \method & 50.2 & \textbf{82.7} & 58.4 & \textbf{42.1} & \textbf{59.9} & \textbf{42.7} & \textbf{73.9} & 49.3 & 32.6 & \textbf{51.3} \\ 
        \bottomrule
    \end{tabular}
    \caption{Main results on Claude 3.5 Sonnet, Gemini 1.5 Pro, Mistral-Large, and Mistral-Nemo under zero-shot setting, showing the accuracy of benchmarked alternatives vs. \method. Best scores are in bold. Note that USC consumes approximately three times more tokens than other RAG methods, and is not directly comparable.
    }
    \label{tab:main}
\end{table*}
\footnotetext{We observe a high refusal rate in RobustRAG for Gemini.}

We evaluate the effectiveness of \method on overcoming imperfect retrieval augmentation and addressing knowledge conflicts. In this section, we introduce the experiment setting~(\Secref{sec:experiment_setting}), compare the performance of \method with various baselines on diverse datasets~(\Secref{sec:experiment_results}), and provide in-depth analyses (\Secref{sec:experiment_analysis}).

\subsection{Experimental Settings}
\label{sec:experiment_setting}

\paragraph{Datasets and metrics.} 
We consider datasets encompass general questions, domain-specific questions, long-tail questions, as well as both short-form and long-form formats, following prior work~\cite{xiang2024certifiably,wei2024instructrag}. 
On NQ, TriviaQA, BioASQ, and PopQA, we provide 10  passages collected with Google Search from the Web for each instance. 
For long-form QA, we use ASQA \cite{stelmakh2022asqa}.
We also evaluate on RGB \citep{chen2024benchmarking}. We choose the English subset (refined version) focusing on noise robustness. For each instance, we select five top negative passages to form a worst-case scenario.
Following prior work, we report the accuracy by string match. More details are in \Appref{sec:benchmark_data}. 

\paragraph{Models and General Settings.}
We conduct experiments on advanced proprietary and open-source LLMs of different scales, including Claude 3.5 Sonnet~(\texttt{claude-3-5-sonnet@20240620}),\footnote{https://www.anthropic.com/news/claude-3-5-sonnet} Gemini 1.5 Pro~(\texttt{gemini-1.5-pro-002}),\footnote{https://deepmind.google/technologies/gemini/pro/} Mistral-Large~(\texttt{128B; version 2407}), and Mistral-Nemo~(\texttt{12B; version 2407}). The generation temperature is set to 0 and the maximum output tokens is set to 1,024. 
All experiments are under the zero-shot setting for controlled evaluation.

\paragraph{Baselines.} 
We compare \method with various RAG methods designed for enhanced robustness.
\textit{USC} \citep{chen2024universal} is a self-consistency method that samples multiple LLM responses and aggregates the answers. 
It provides a reference of naive improvements using additional API calls. 
\textit{Genread} \citep{yu2023generate} augments retrieved passages with LLM-generated passages without explicit consolidation process. 
\textit{RobustRAG} \citep{xiang2024certifiably} aggregates answers from independent passages to provide certifiable robustness. We use the best-performing keyword aggregation variant.
\textit{InstructRAG} \citep{wei2024instructrag} instructs the LLM to provide a rationale connecting the answer with information in passages. For a fair comparison, no training is applied.
\textit{Self-Route} \citep{xu2024retrieval} adaptively switches between LLMs with and without RAG.\footnote{The original Self-Route switches between RAG and long-context LLMs, while our implementation switches between RAG and No RAG according to our problem formulation.} It provides a reference of switching between LLMs' internal and external knowledge.

\paragraph{Implementation Details.}
The prompt templates for \method can be found in \Appref{sec:appendix_prompt}. 
By default, we set $t=1$ and $\hat{m}=1$ to limit the number of additional tokens used. 
Results with larger $t$ and $\hat m$ are discussed in \Secref{sec:experiment_analysis}.

\begin{figure}[t]
  \begin{center}
    \includegraphics[width=0.7\columnwidth]{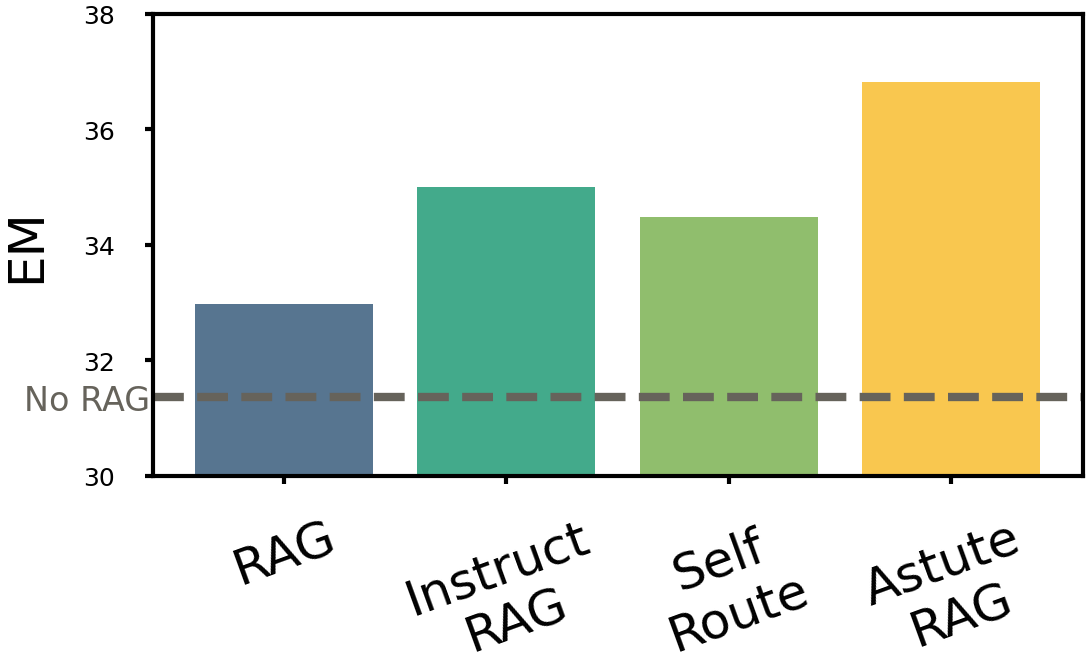}
  \end{center}
  \caption{Performance on ASQA.}\label{fig:asqa}
\end{figure}

\begin{figure}[t]
  \begin{center}
    \includegraphics[width=0.7\columnwidth]{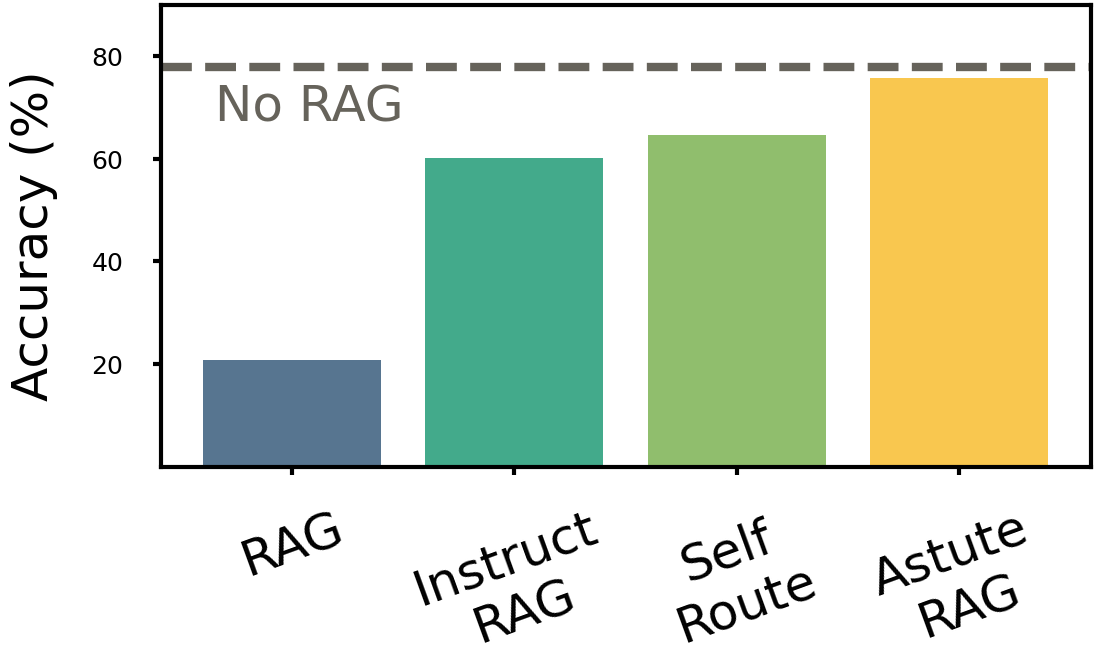}
  \end{center}
  \caption{Worst-case performance of Claude on RGB. \method reaches a performance close to No RAG, while other RAG systems are far behind.}\label{fig:rgb}
\end{figure}

\subsection{Main Results}
\label{sec:experiment_results}

\paragraph{Performance under real-world retrieval.}
\Tabref{tab:main} presents the results with real-world retrieval augmentation of various LLMs. 
We find that retrieved passages might not always bring benefits -- on NQ and TriviaQA, RAG performance lags behind No RAG for advanced LLMs. We attribute this questions being covered by the LLM's internal knowledge and the noise in retrieval results misleading the LLM. 
In contrast, on BioASQ and PopQA, which focus on domain-specific and `long-tail' questions, RAG significantly improves the LLM performance. 
Due to imperfect retrieval augmentation, however, the absolute performance still remains to be unsatisfactory.
Among all baselines, no single method consistently outperforms others across all datasets and LLMs. This observation highlights these baselines being tailored to distinct settings and not being universally applicable.
Overall, InstructRAG and Self-Route demonstrate relatively superior performance among other alternatives.
\method consistently outperforms baselines across all LLMs in terms of overall accuracy. The relative improvement compared to the best baseline is 6.85\% for Claude and 4.13\% for Gemini, with the improvements in domain-specific questions being much higher. These highlight the effectiveness of \method in overcoming imperfect retrieval augmentation and knowledge conflicts. 
Additionally, we observe consistent improvements on the open-source Mistral models. The results demonstrate that \method generalizes well to LLMs of smaller sizes.

\begin{figure}[t]
  \begin{center}
    \includegraphics[width=0.9\columnwidth]{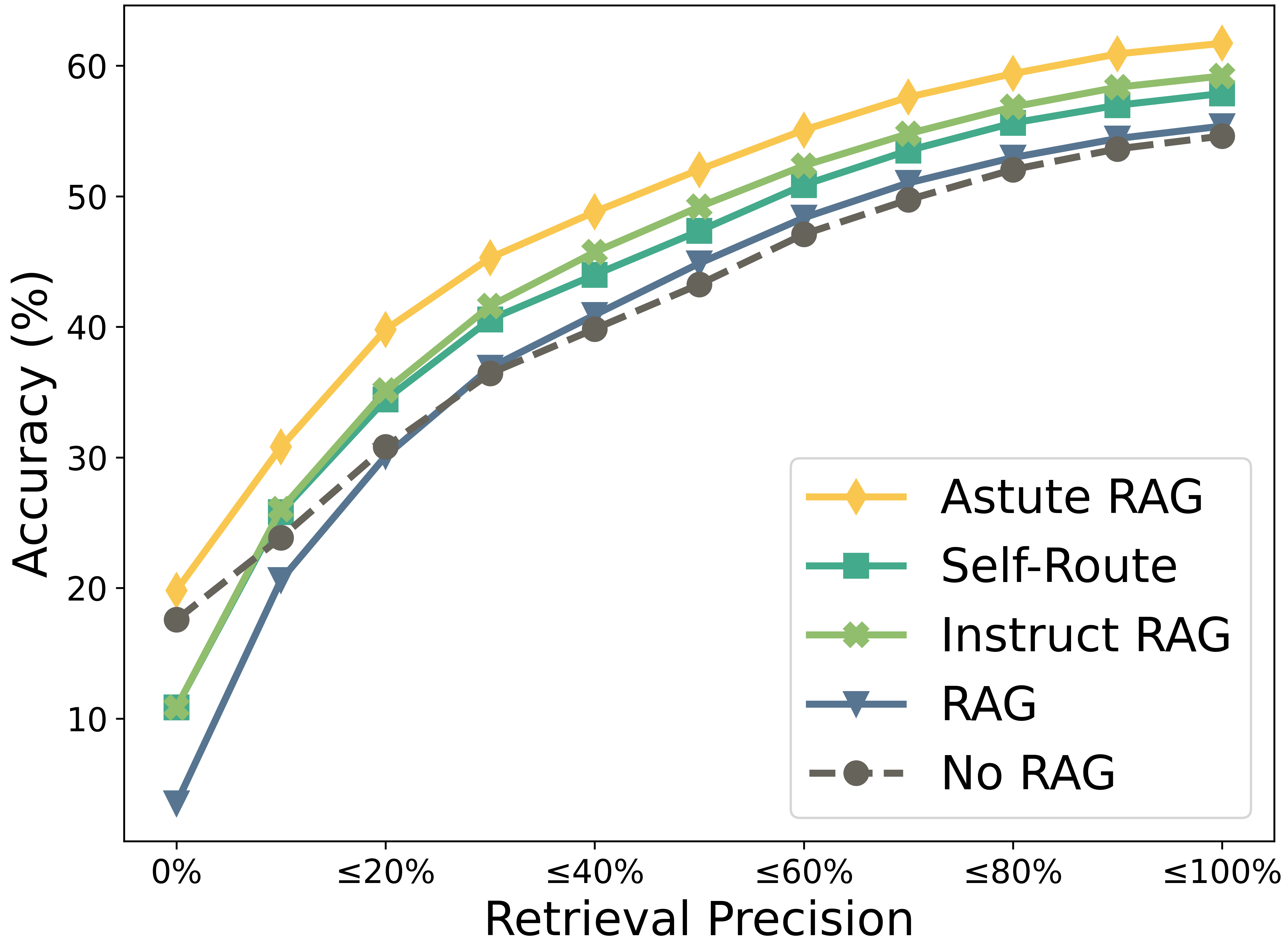}
  \end{center}
  \caption{Performance across different retrieval precision buckets. \method is consistently better.}
  \label{fig:perf_by_retrieval}
\end{figure}

\paragraph{Performance on long-form QA.}
We conduct additional experiments on the long-form QA dataset, ASQA. \Figref{fig:asqa} demonstrates that \method consistently achieves significant improvements, reinforcing its effectiveness across diverse scenarios.

\paragraph{Worst-case performance on RGB.}
\Figref{fig:rgb} presents the results under the worst-case setting on RGB where all retrieved documents are negative, to demonstrate robustness.
The performance gap between RAG and No RAG exceeds 50 points, highlighting the detrimental impact of imperfect retrieval results and emphasizing the importance of providing robust safeguards against worst-case scenarios.
While the baseline RAG methods outperform the original RAG, they still obviously fall behind `No RAG'. \method is the only RAG method that reaches a performance close to `No RAG', further supporting its effectiveness in addressing imperfect retrieval augmentation.

\subsection{Analyses}
\label{sec:experiment_analysis}

We conduct in-depth analyses using Claude following the setting of \Tabref{tab:main}.

\paragraph{The impact of retrieval precision.}
As shown in \Figref{fig:perf_by_retrieval}, \method achieves consistently better performance across different retrieval precision regimes, indicating its effectiveness in improving RAG trustworthiness in broad scenarios. Notably, \method does not sacrifice performance gain under high retrieval quality in exchange for improvement under low retrieval quality.
When the retrieval quality is extremely low (close to zero precision), all other RAG variants underperform the 'No RAG' baseline, except for \method.

\begin{figure}[t]
    \centering
    \includegraphics[width=0.8\columnwidth]{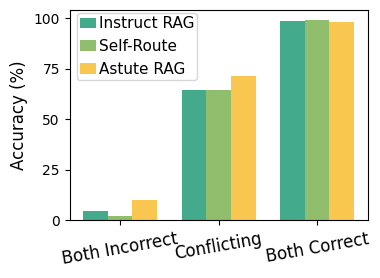}
    \caption{Performance on conflicting and consistent instances between No RAG and RAG.}
    \label{fig:perf_by_conflicts}
\end{figure}

\paragraph{Addressing knowledge conflicts.}
We split our collected data into three subset according to the answers with and without RAG: the answers from two can be (i) both correct, (ii) both incorrect, or (iii) conflicting with one being correct. 
The results are shown in \Figref{fig:perf_by_conflicts}.
On the conflicting subset, \method successfully chooses the correct answer in approximately 80\% of cases, being the most effective one in addressing knowledge conflicts.
Notably, \method even brings performance improvement on the subset where neither internal nor external knowledge alone leads to the correct answer. This indicates that \method can effectively combine partially-correct information from LLM-internal and external knowledge.

\begin{figure}[t]
\centering
\includegraphics[width=\columnwidth]{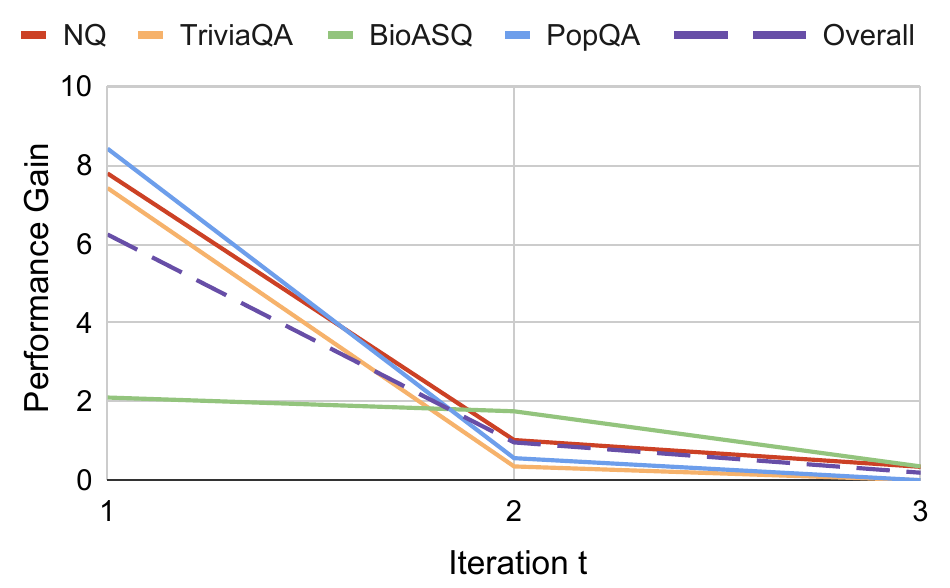}
\caption{Accuracy improvement when increasing $t$.}
\label{fig:iteration}
\end{figure}

\paragraph{Benefits of more consolidation iteration.}
For efficiency, we employ a single iteration of knowledge consolidation in our main experiments. However, incorporating multiple iterations has the potential to further enhance model performance as shown in \Figref{fig:iteration}. The magnitude of this improvement diminishes as $t$ increases, indicating that the knowledge has been better presented and less improvement space left after each iteration.

\begin{figure}[t]
\centering
\includegraphics[width=\columnwidth]{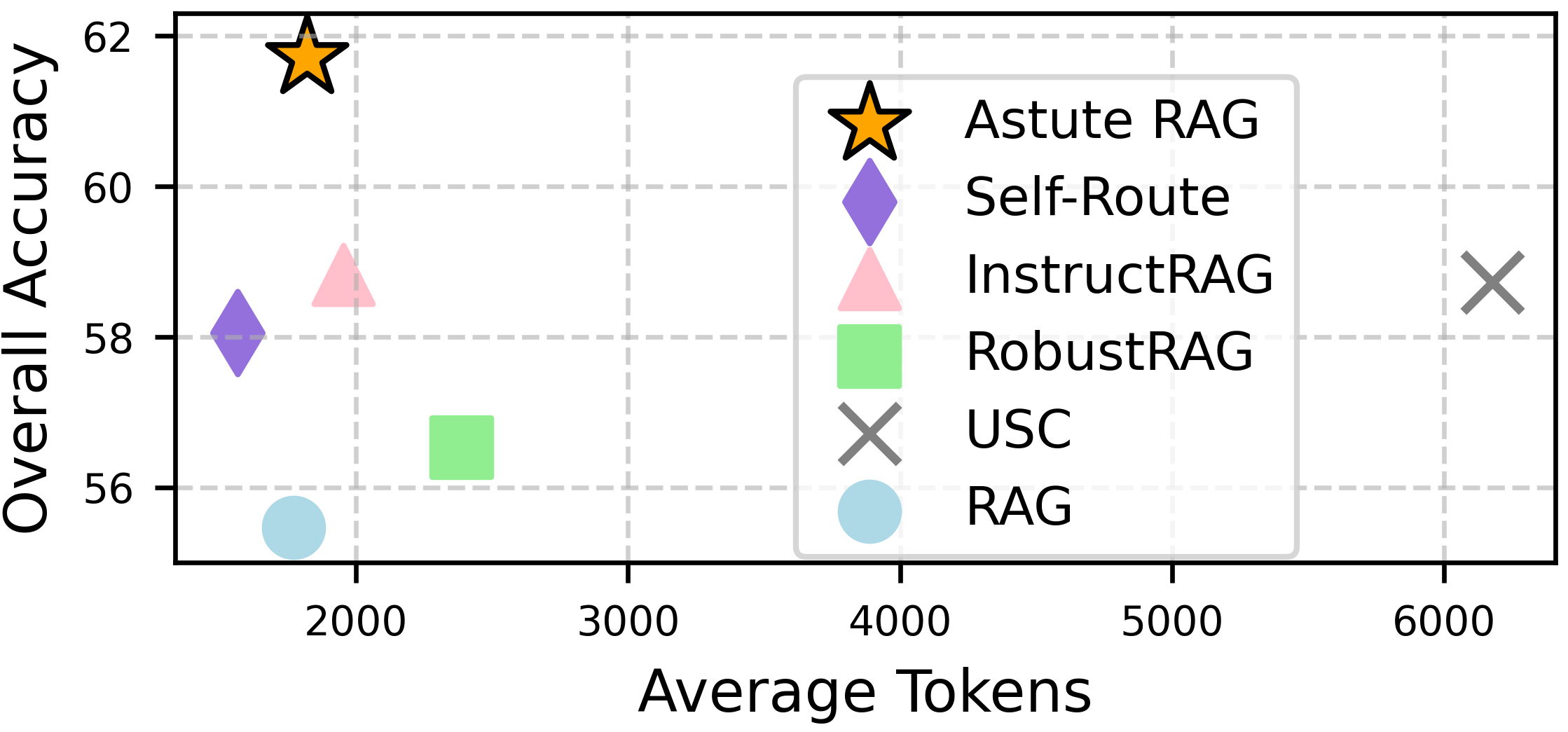}
\caption{Efficiency in terms of tokens consumed.}
\label{fig:token}
\end{figure}

\paragraph{Efficiency in tokens consumed and API calls.}
As a proxy to overall prediction cost and latency, we present the average number of tokens and API calls used per instance in \Figref{fig:token} and  \Figref{fig:api_call}. \method incurs only a marginal cost increase, <5\%, while delivering substantial improvement, >11\%, compared to the RAG baseline.

\paragraph{Effectiveness of adaptive generation.}
The results in \Tabref{tab:m} illustrate the model's performance when varying the maximum number of passages generated. The design of adaptive generation has been effectively reflected, as the number of generated passages is dynamically adjusted leading to $m < \hat{m}$.
Notably, the number of generated passages can be controlled by $\hat{m}$, and results show that the system does not generate passages excessively.

\paragraph{Impact of Source-Awareness.}
To evaluate the impact of source-awareness, we conducted an ablation study where source labels (\textit{own memory} and \textit{external retrieval}) were removed during the consolidation and answer generation process. The results are shown in \Tabref{tab:source-awareness-ablation}.
The comparison shows that providing source information (\method) leads to better performance overall compared to omitting it (\textit{\method (No Source)}), particularly on NQ, PopQA, and BioASQ, suggesting that awareness of information origin aids the consolidation process.

\paragraph{Accuracy of intermediate steps.}
To investigate the performance of intermediate steps, including knowledge consolidation and confidence assignment, we use LLM-as-a-judge with the instruction in \Appref{sec:appendix_prompt}. Our experimental results show that the accuracy for knowledge consolidation is 98.2\%, and for confidence assignment, it is 95.0\%. These results demonstrate the effectiveness of the proposed framework in the intermediate stages.

\paragraph{Qualitative examples.}
In \Figref{fig:qual}, we present two representative examples showing the intermediate outputs of \method. In the first example, LLM without RAG generates a wrong answer, while RAG returns a correct answer. \method successfully identified the incorrect information in its generated passage and an external passage, avoiding confirmation bias \cite{tan2024blinded}. In the second example, LLM is correct but RAG is incorrect due to imperfect retrieval. \method detected the correct answer from imperfect context leveraging internal knowledge.

\section{Conclusion}

We investigate the impact of imperfect retrieval on the performance of RAG systems and identify knowledge conflicts as a key challenge. To address this, we introduce \method, a novel approach that leverages the internal knowledge of LLMs and iteratively refines the generated responses by consolidating internal and external knowledge in a source way. We demonstrate the effectiveness of \method in mitigating the negative effects of imperfect retrieval and improving the robustness of RAG, particularly in challenging scenarios with unreliable external sources.

\section*{Acknowledgement}

We would like to thank Jinsung Yoon for valuable discussions and insights that helped to improve this paper. %
We would also like to thank all other colleagues from Google Cloud AI Research for their valuable feedback.

\section*{Limitations}

\method's effectiveness hinges on the capabilities of advanced LLMs with strong instruction-following and reasoning abilities, hence potentially more limited applicability with less sophisticated LLMs. 
As an important future direction,  extending the experimental setup to include longer inputs would be important, where the challenges of imperfect retrieval and knowledge conflicts may be even more pronounced.

\bibliography{custom}

\clearpage
\appendix
\appendix

\begin{table*}[h]
    \centering
    \small
    \begin{tabular}{l|llllll}
    \toprule
        ~ & NQ & TriviaQA & BioASQ & PopQA & Overall & $m$ \\ \midrule
        $\hat m$=1 & 52.20 & 84.10 & 60.14 & 44.38 & 61.71 & 0.69 \\ 
        $\hat m$=2 & 52.20 & 85.16 & 60.84 & 43.26 & 62.00 & 1.24 \\ \bottomrule
    \end{tabular}
    \caption{Performance and averge number of generaed passages using different $\hat m$.}
    \label{tab:m}
\end{table*}

\begin{figure*}[h]
\centering
\includegraphics[width=\columnwidth]{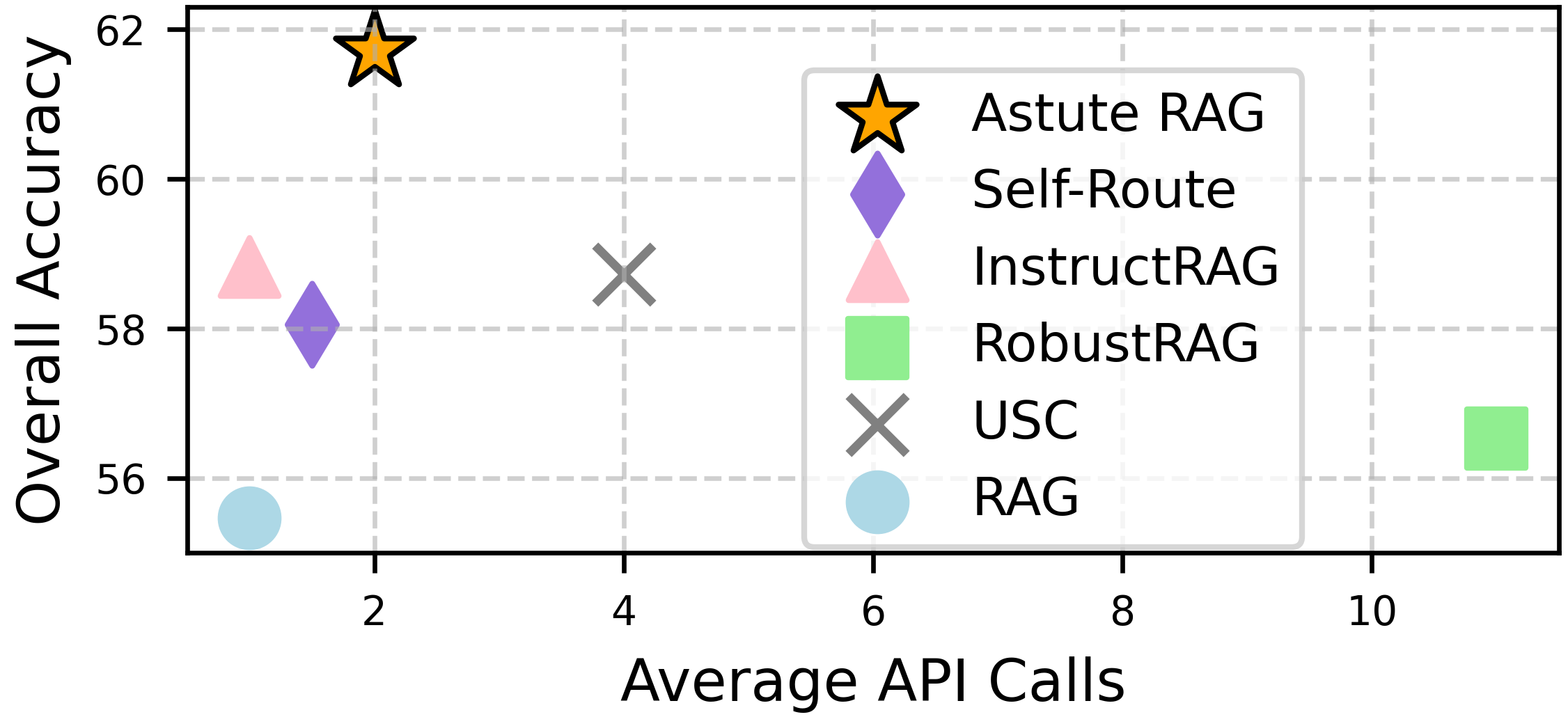}
\caption{Efficiency in terms of API calls.}
\label{fig:api_call}
\end{figure*}

\begin{table*}[h]
\small
\centering
\begin{tabular}{lccccc}
\toprule
Model & NQ & TriviaQA & BioASQ & PopQA & Overall \\
\midrule
\method             & \textbf{50.2} & 81.6 & \textbf{58.0} & \textbf{40.5} & \textbf{59.2} \\
\method (No Source) & 48.1 & 82.3 & 57.7 & 39.9 & 58.6 \\
\bottomrule
\end{tabular}
\caption{Ablation study on source-awareness.}
\label{tab:source-awareness-ablation}
\end{table*}

\begin{figure*}[h]
\centering
\includegraphics[width=0.95\textwidth]{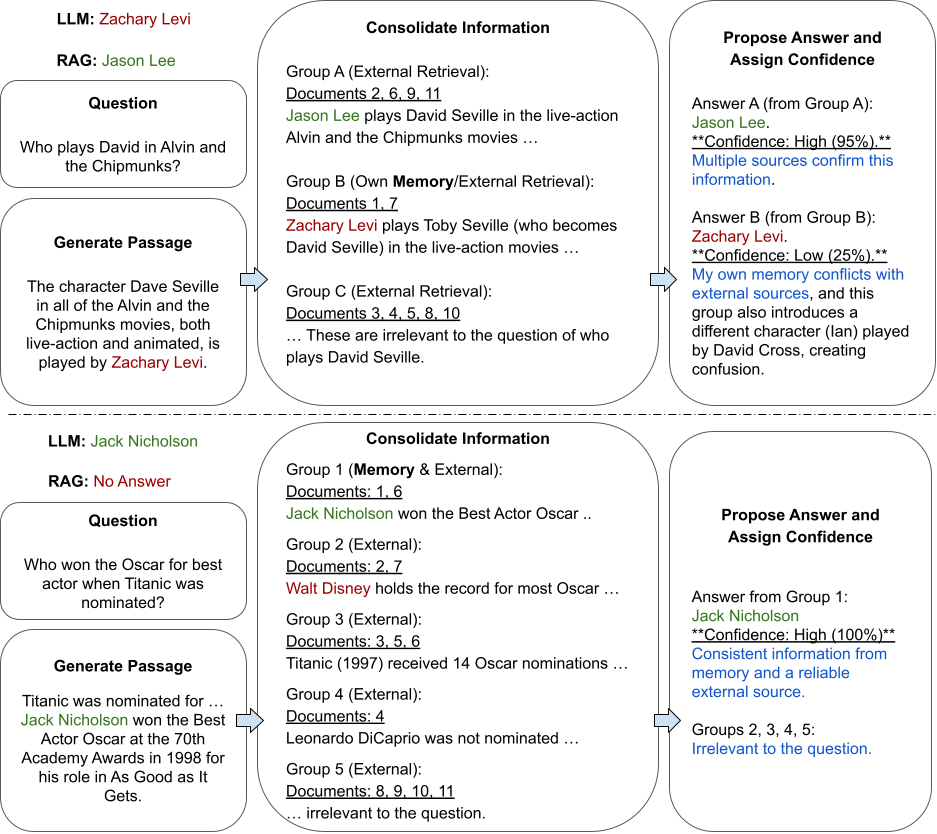}
\caption{Qualitative examples. \textit{Top:} \method identified the error in internal knowledge (i.e., generated passage) by confirming with external sources. \textit{Bottom:} \method detected the correct answer from imperfect retrieval by checking with its internal knowledge. Standard RAG does not provide an answer because the retrieved passages are too noisy.}
\label{fig:qual}
\end{figure*}

\clearpage

\section{Prompt Template for \method}
\label{sec:appendix_prompt}

\begin{multicols}{2}
\begin{tcolorbox}[colback=gray!10, colframe=black, title=Prompt for Adaptive Passage Generation ($p_{gen}$), width=\textwidth]
Generate a document that provides accurate and relevant information to answer the given question. If the information is unclear or uncertain, explicitly state 'I don't know' to avoid any hallucinations.  
\\ \\
Question: \{question\}
Document:
\end{tcolorbox}
\end{multicols}

\begin{multicols}{2}
\begin{tcolorbox}[colback=gray!10, colframe=black, title=Prompt for Iterative Knowledge Consolidation ($p_{con}$), width=\textwidth]
Task: Consolidate information from both your own memorized documents and externally retrieved documents in response to the given question.
\\ \\
* For documents that provide consistent information, cluster them together and summarize the key details into a single, concise document.

* For documents with conflicting information, separate them into distinct documents, ensuring each captures the unique perspective or data.

* Exclude any information irrelevant to the query.

For each new document created, clearly indicate:

* Whether the source was from memory or an external retrieval.

* The original document numbers for transparency.
\\ \\
Initial Context: \{context\}

Last Context: \{context\}

Question: \{question\}

New Context:
\end{tcolorbox}
\end{multicols}

\clearpage

\begin{multicols}{2}
\begin{tcolorbox}[colback=gray!10, colframe=black, title=Prompt for Knowledge Consolidation and Answer Finalization ($p_{ans}$), width=\textwidth]
Task: Answer a given question using the consolidated information from both your own memorized documents and externally retrieved documents.
\\ \\
Step 1: Consolidate information

* For documents that provide consistent information, cluster them together and summarize the key details into a single, concise document.

* For documents with conflicting information, separate them into distinct documents, ensuring each captures the unique perspective or data.

* Exclude any information irrelevant to the query.

For each new document created, clearly indicate:

* Whether the source was from memory or an external retrieval.

* The original document numbers for transparency.
\\ \\
Step 2: Propose Answers and Assign Confidence

For each group of documents, propose a possible answer and assign a confidence score based on the credibility and agreement of the information.
\\ \\
Step 3: Select the Final Answer

After evaluating all groups, select the most accurate and well-supported answer.

Highlight your exact answer within $<$ANSWER$>$ your answer $<$/ANSWER$>$.
\\ \\
Initial Context: \{context\_init\}

[Consolidated Context: \{context\}] \# optional

Question: \{question\}

Answer:
\end{tcolorbox}
\end{multicols}

\begin{multicols}{2}
\begin{tcolorbox}[colback=gray!10, colframe=black, title=Prompt for Intermediate Step Evaluation, width=\textwidth]
**Task:** You are provided with the following: 

1. A question.  

2. The correct answer.  

3. The input context.  

4. The model's response, which contains:

   - Consolidated context.  
   
   - Confidence scores for candidate answers.  

Your task is to:  

- Evaluate the **quality of the consolidated context** in the model's response and provide a label: `$<$consolidation$>$ correct $<$/consolidation$>$' or `$<$consolidation$>$ incorrect $<$/consolidation$>$'. This evaluation is only about whether the consolidation is correct given the input context. 

- Evaluate the **accuracy of the confidence score** (whether it aligns with the confidence of the supporting context) and provide a label: `$<$confidence$>$ correct $<$/confidence$>$' or `$<$confidence$>$ incorrect $<$/confidence$>$'. The evaluation is only based on the consolidated context.  

Note that correct consolidation and confidence do not necessarily indicate the correct answer.

Question: 
\{query\}

Correct Answer: 
\{answer\}

Input Context: 
\{input\}

Model Response: 
\{response\}

Evaluation:
\end{tcolorbox}
\end{multicols}

\clearpage

\section{Data Collection}
\label{sec:benchmark_data}

Encompassing a \textit{diverse} range of \textit{natural} questions, our benchmark consists of \textit{realistic} retrieval results with Google Search\footnote{\url{https://developers.google.com/custom-search/v1/overview}} as the retriever and the Web as the corpus. Notably, we do not select questions or annotate answers based on the retrieval results. This setting allows us to analyze the severity of imperfect retrieval in real-world RAG. It distinguishes our benchmark from previous ones that employ synthetic retrieval corruptions or that unintentionally reduce the frequency of imperfect retrieval with biased construction protocols \citep{chen2024benchmarking,yang2024crag}. Overall, our benchmark contains 1,042 short-form question-answer pairs, each paired with 10 retrieved passages. 
When collecting the passages, we retrieve the top 30 results and select the first 10 accessible websites. From each retrieved website, we extract the paragraph corresponding to the snippet provided in the search results as the retrieved passage. Retrieved results might contain natural noise with irrelevant or misleading information. We do not consider enhancements to the retrieval side, such as query rewriting, as such enhancements are typically already incorporated into commercial information retrieval systems. 
All of these datasets are short-form QA. Following previous work \citep{xiang2024certifiably,wei2024instructrag,mallen2023not}, a model response is considered correct if it contains the ground-truth answer. To enhance evaluation reliability, we prompt LLMs to enclose the exact answer within special tokens, extracting them as the final responses. 

\paragraph{Question-answer pairs.} 
We consider question-answer pairs from four datasets of different properties spanning across general questions, domain-specific questions, and long-tail questions. NQ \citep{kwiatkowski2019natural} and TriviaQA \citep{joshi2017triviaqa} are two widely-studied question-answering (QA) datasets in general domains. BioASQ \citep{tsatsaronis2015overview} is from biomedical domain that has demonstrated significant benefits from RAG when general-purpose LLMs are considered. PopQA \citep{mallen2023not} focuses on long-tail knowledge and has been shown to be challenging for even advanced LLMs to solve without external knowledge. All these datasets contain questions with short-form answers and most of them list all valid answer variants. This format can support automatic verification of answer appearance in retrieved passages and model responses, leading to more precise evaluations.

\paragraph{Retrieval process.}
For each question in our benchmark, we query Google Search to retrieve the top 30 results and select the first 10 accessible websites. From each retrieved website, we extract the paragraph corresponding to the snippet provided in Google Search results as the retrieved passage. We do not consider enhancements to the retrieval side, such as query rewriting, as such enhancements are typically already incorporated into commercial information retrieval systems.

\section{Comparison with Answer Refinement}

In \Tabref{tab:refinement-routing-comparison}, we further compare \method with \textit{Answer Refinement}, where the LLM is prompted to refine its initial answer by reconsidering external context. Notably, this baseline performs nearly identically to \textit{Self-Route}, which is expected, because both approaches rely on the model itself to determine whether to revise its initial answer based on external knowledge. The comparison underscores that simply enabling LLMs to self-correct does not yield significant improvements beyond existing routing strategies. In contrast, \method continues to outperform all baselines across datasets, reinforcing the benefit of explicitly structured consolidation mechanisms.

\begin{table*}[h]
\small
\centering
\begin{tabular}{lccccc}
\toprule
Model & NQ & TriviaQA & BioASQ & PopQA & Overall \\
\midrule
Self-Route        & 47.5 & 79.9 & 58.0 & 38.2 & 57.6 \\
Answer Refinement & 47.1 & 79.9 & 58.0 & 38.2 & 57.5 \\
\method        & 50.2 & 81.6 & 58.0 & 40.5 & 59.2 \\
\bottomrule
\end{tabular}
\caption{Comparison of Astute RAG with routing and refinement-based baselines on Gemini.}
\label{tab:refinement-routing-comparison}
\end{table*}

\section{Comparison with Context Filtering}

To isolate the effect of identifying irrelevant information separate from consolidation, we conducted an ablation study introducing two \textit{Context Filtering} baselines. The first baseline filters only the retrieved documents, while the second filters both retrieved and generated documents prior to answer generation, both without applying the consolidation step. As shown in \Tabref{tab:context-filtering}, while context filtering improves performance over the basic RAG baseline, it falls short of the performance achieved by \method. This supports our hypothesis that consolidating diverse information, including consistent, conflicting, and relevant content, rather than merely filtering out irrelevant parts, is critical to the performance gains observed with \method.

\begin{table*}[h]
\small
\centering
\begin{tabular}{lccccc}
\toprule
Model & NQ & TriviaQA & BioASQ & PopQA & Overall \\
\midrule
RAG                                         & 42.7 & 76.0 & 55.2 & 33.7 & 53.7 \\
Context Filtering (retrieved)              & 43.7 & 77.0 & 57.0 & 34.3 & 54.8 \\
Context Filtering (generated and retrieved) & 49.2 & 79.2 & 56.6 & 40.4 & 57.9 \\
Astute RAG                                  & 50.2 & 81.6 & 58.0 & 40.5 & 59.2 \\
\bottomrule
\end{tabular}
\caption{Comparison with context filtering baselines.}
\label{tab:context-filtering}
\end{table*}

\section{Comparison with Context Compression}
Context compression \cite{wang2023learning,yoon2024compact} is also a related direction.
We further conduct experiments comparing our method with CompAct \cite{yoon2024compact}. The results in \Tabref{tab:ef_compare_claude} and \Tabref{tab:ef_compare_gemini} show that context compression is ineffective in handling the challenges of imperfect context and knowledge conflicts. Notably, it even performs worse than the No RAG and RAG baselines in this context. The primary reason for this underperformance lies in the limitations of context compression. It struggles to effectively identify incorrect information when there are conflicts in context and often removes or reduces the appearance of helpful information in the process. This reinforces the importance of our approach, which does not rely solely on compression but instead integrates both internal and external knowledge while handling conflicts in a more nuanced manner.

\begin{table*}[!ht]
    \centering
    \small
    \begin{tabular}{lccccc}
    \toprule
        Method & NQ & TriviaQA & BioASQ & PopQA & Overall \\ \midrule
        No RAG & 47.1 & 82.0 & 50.4 & 29.8 & 54.5 \\ 
        RAG & 44.4 & 76.7 & 58.0 & 36.0 & 55.5 \\ 
        CompAct & 38.6 & 68.9 & 49.3 & 30.3 & 48.4 \\ 
        \method & 52.2 & 84.1 & 60.1 & 44.4 & 61.7 \\ \bottomrule
    \end{tabular}
    \caption{Comparison with context compression on Claude.}
    \label{tab:ef_compare_claude}
\end{table*}

\begin{table*}[h]
    \centering
    \small
    \begin{tabular}{lccccc}
    \toprule
        Method & NQ & TriviaQA & BioASQ & PopQA & Overall \\ \midrule
        No RAG & 44.8 & 80.2 & 45.8 & 25.3 & 51.3 \\ 
        RAG & 42.7 & 76.0 & 55.2 & 33.7 & 53.7 \\ 
        CompAct & 35.3 & 65.0 & 47.6 & 30.9 & 46.0 \\ 
        \method & 50.2 & 81.6 & 58.0 & 40.5 & 59.2 \\ \bottomrule
    \end{tabular}
    \caption{Comparison with context compression on Gemini.}
    \label{tab:ef_compare_gemini}
\end{table*}

\section{Influence of passage ordering.}
We apply different ordering strategies \citep{alessio2024improving} on RAG and \method. As shown in \Tabref{tab:ordering}, we find that the improvement with \method is significantly larger than the gap between different ordering strategies. Moreover, the consolidation process makes \method less sensitive to it.

\begin{table*}[h]
    \centering
    \small
    \begin{tabular}{ll|ccccc}
    \toprule
        Method & Ordering Strategy & NQ & TriviaQA & BioASQ & PopQA & Overall \\ \midrule
        RAG & Random & 43.39 & 76.33 & 56.99 & 34.83 & 54.61 \\ 
        ~ & Ascending & 43.05 & 75.62 & 57.69 & 34.83 & 54.51 \\ 
        ~ & Descending & 44.41 & 76.68 & 58.04 & 35.96 & 55.47 \\ 
        ~ & Ping-pong Descending Top-to-bottom & 44.75 & 77.39 & 57.69 & 35.96 & 55.66 \\ 
        ~ & Ping-pong Descending Bottom-to-top & 44.41 & 75.62 & 58.04 & 35.96 & 55.18 \\ \midrule
        AstuteRAG & Random & 51.86 & 84.81 & 61.19 & 41.57 & 61.61 \\ 
        ~ & Ascending & 51.86 & 85.51 & 59.79 & 42.13 & 61.52 \\ 
        ~ & Descending & 52.20 & 84.10 & 60.14 & 44.38 & 61.71 \\ 
        ~ & Ping-pong Descending Top-to-bottom & 52.20 & 84.45 & 59.09 & 43.82 & 61.42 \\ 
        ~ & Ping-pong Descending Bottom-to-top & 51.19 & 85.16 & 61.54 & 43.82 & 62.00 \\ \bottomrule
    \end{tabular}
    \caption{Performance by Ordering Strategies.}
    \label{tab:ordering}
\end{table*}

\end{document}